\documentclass[10pt,twocolumn,letterpaper]{article}

\usepackage{cvpr}              

\usepackage{xcolor,pifont}
\usepackage{colortbl}
\usepackage{wrapfig}
\usepackage{overpic}
\usepackage{graphicx}
\usepackage{amsmath}
\usepackage{subcaption}
\usepackage{multirow}
\usepackage{tcolorbox}
\usepackage[font=small,labelfont=bf]{caption}

\newcommand{\cmark}{\ding{52}}%
\newcommand{\xmark}{\ding{55}}%

\usepackage{xcolor}
\definecolor{citecolor}{RGB}{34,139,34}
\definecolor{lightred}{RGB}{255,100,100}
\definecolor{cell_bisque}{rgb}{1.0, 0.89, 0.77}
\definecolor{cell_blond}{rgb}{0.98, 0.94, 0.75}
\definecolor{cell_blue}{RGB}{155, 187, 228}
\definecolor{princetonorange}{rgb}{1.0, 0.56, 0.0}
\definecolor{pinkpearl}{rgb}{0.91, 0.67, 0.81}
\definecolor{mossgreen}{rgb}{0.68, 0.87, 0.68}
\definecolor{cadmiumgreen}{rgb}{0.0, 0.42, 0.24}
\definecolor{brightmaroon}{rgb}{0.76, 0.13, 0.28}
\definecolor{calpolypomonagreen}{rgb}{0.12, 0.35, 0.17}
\definecolor{darkseagreen}{rgb}{0.56, 0.74, 0.56}
\definecolor{armygreen}{rgb}{0.29, 0.33, 0.13}
\definecolor{azure}{rgb}{0.0, 0.5, 1.0}
\definecolor{denim}{rgb}{0.08, 0.38, 0.84}

\newcommand{\Paragraph}[1]{\vspace{-0mm}\noindent\textbf{#1.}\hspace{0mm}}
\newcommand{\Section}[1]{\vspace{-1mm} \section{#1} \vspace{-0mm}}
\newcommand{\SubSection}[1]{\vspace{-0mm} \subsection{#1} \vspace{-0mm}}
\newcommand{\SubSubSection}[1]{\vspace{-0mm} \subsubsection{#1} \vspace{-0mm}}

\definecolor{cvprblue}{rgb}{0.21,0.49,0.74}
\definecolor{linkcolor}{rgb}{0.93, 0.11, 0.14}
\definecolor{citecolor}{rgb}{0, 113, 188}
\usepackage[pagebackref,breaklinks,colorlinks,linkcolor=linkcolor,citecolor=cvprblue]{hyperref}

\def\paperID{13948} 
\def\confName{CVPR}
\def\confYear{2025}

\title{Rethinking Vision-Language Model in Face Forensics: \\
Multi-Modal Interpretable Forged Face Detector
}

\begin{document}

\author{Xiao Guo$^{1}$, Xiufeng Song$^{2}$, Yue Zhang$^{1}$, Xiaohong Liu$^{2}$, Xiaoming Liu$^{1}$\\
$^{1}$ Michigan State University $^{2}$ Shanghai Jiao Tong University \\
{\tt\small \{guoxia11, zhan1624, liuxm\}@msu.edu, \tt\small \{akikaze,xiaohongliu\}@sjtu.edu.cn}
}

\maketitle
\begin{abstract}
Deepfake detection is a long-established research topic vital for mitigating the spread of malicious misinformation.
Unlike prior methods that provide either binary classification results or textual explanations separately, we introduce a novel method capable of generating both simultaneously.
Our method harnesses the multi-modal learning capability of the pre-trained CLIP and the unprecedented interpretability of large language models (LLMs) to enhance both the generalization and explainability of deepfake detection. 
Specifically, we introduce a multi-modal face forgery detector (M2F2-Det) that employs tailored face forgery prompt learning, incorporating the pre-trained CLIP to improve generalization to unseen forgeries.
Also, M2F2-Det incorporates an LLM to provide detailed textual explanations of its detection decisions, enhancing interpretability by bridging the gap between natural language and subtle cues of facial forgeries.
Empirically, we evaluate M2F2-Det on both detection and explanation generation tasks, where it achieves state-of-the-art performance, demonstrating its effectiveness in identifying and explaining diverse forgeries.
Source code is available at \href{https://github.com/CHELSEA234/M2F2_Det}{link}.
\end{abstract}
\section{Introduction}
Generative Models (GMs)~\cite{goodfellow2014generative, choi2018stargan, karras2019style,rombach2022high,guo2024dense} have demonstrated impressive capabilities in synthesizing highly realistic and visually compelling images. However, they also facilitate the proliferation of AI-generated content (AIGC), like \textit{deepfakes}, raising serious concerns over the spread of deceptive or manipulated facial imagery.
To counter these threats, substantial efforts have been made to develop deepfake detection techniques, including subtle artifacts indentification~\cite{chen2022self,li2020face,liexposing,shiohara2022detecting,zhao2021learning,xiao_hifinet_plusplus}, frequency analysis~\cite{qian2020thinking,gu2022exploiting,liu2021spatial,luo2021generalizing,wang2023dynamic}, disentangling forgery traces via specialized neural networks~\cite{hifi_net_xiaoguo,pscc-net,dong2023implicit,huang2023implicit,yan2023ucf,liang2022exploring,yang2019exposing_uadf,zhao2021multi}, modeling temporal inconsistencies~\cite{haliassos2021lips,wang2023altfreezing,zheng2021exploring,xiufeng_lamma_detection}, among others.

\begin{figure}[t]
    \centering
    \begin{overpic}[width=1\linewidth]{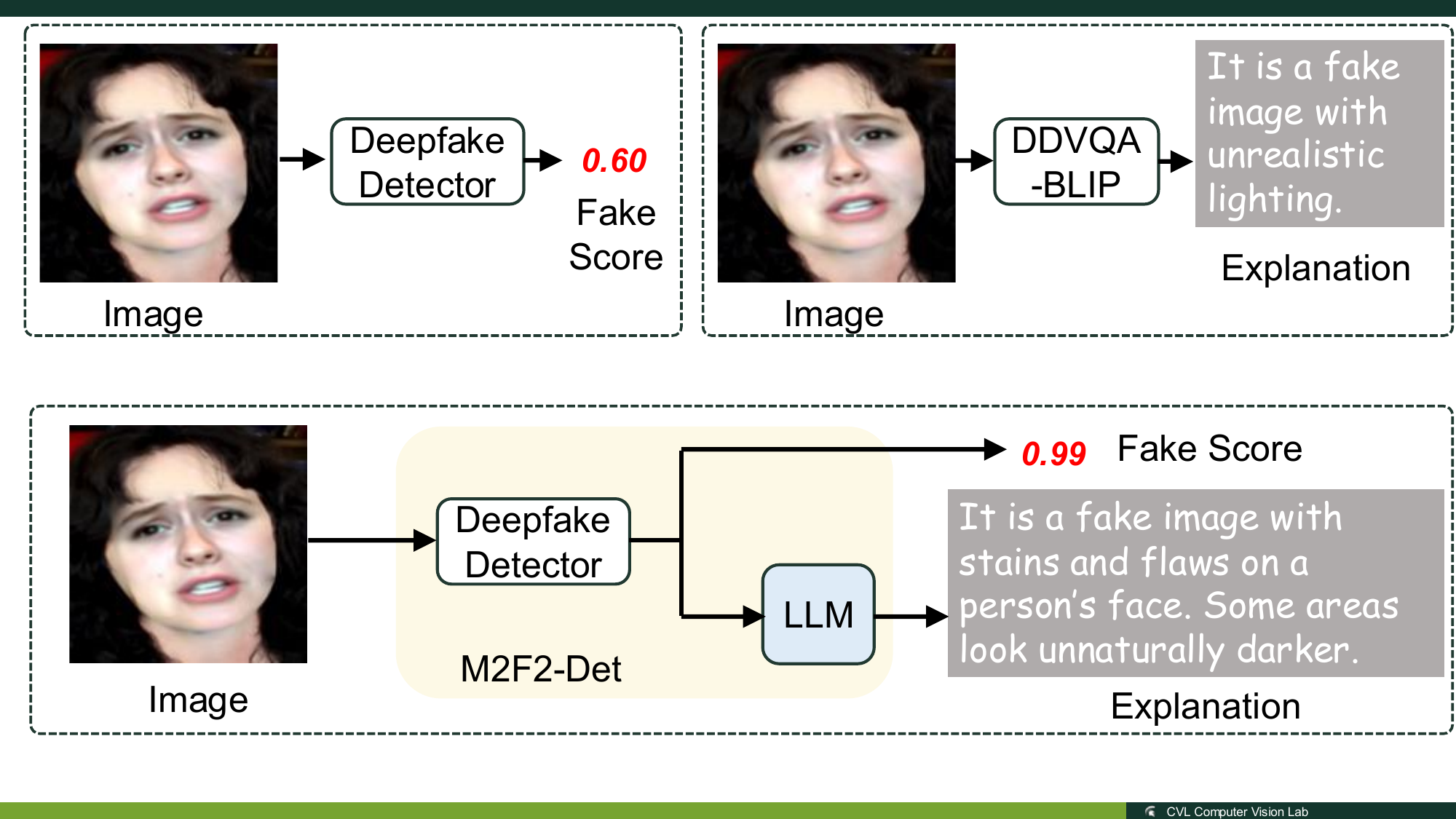}
    \put(25,28){(a)}
    \put(72,28){(b)}
    \put(48,0){(c)} 
    \end{overpic}
    \caption{
    (a) and (b) represent conventional deepfake detectors and DDVQA-BLIP~\cite{zhang2024common}, which take an image as the input and output the fake probability (\textit{e.g.}, score) and textual explanations, respectively. 
    (c) In this work, we propose a multi-modal face forgery detector (M2F2-Det) that produces both fake probability and textual explanations.     
    \vspace{-2mm}
    }
    \label{fig:overall}
\end{figure}

\begin{figure*}[t]
  \centering
    \begin{overpic}[width=1\linewidth]{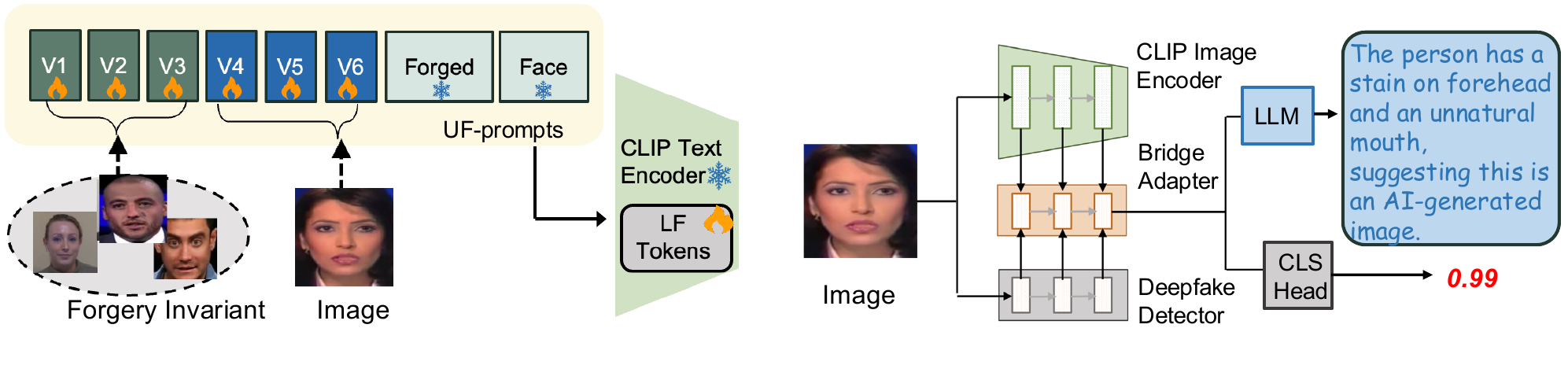}
    \put(23,0){\large(a)}
    \put(71,0){\large(b)}
    \end{overpic}
    \vspace{-3mm}
    \caption{
    (a) Forgery Prompt Learning (FPL) adapts CLIP to deepfake detection by optimizing \textit{UF-prompts} and \textit{layer-wise forgery tokens} (LF tokens).
    UF-prompts consist of three segments: trainable general forgery tokens (\textit{i.e.}, \textcolor{calpolypomonagreen}{$\mathbf{V}_1$}, \textcolor{calpolypomonagreen}{$\mathbf{V}_2$}, and \textcolor{calpolypomonagreen}{$\mathbf{V}_3$}), specific forgery tokens (\textit{i.e.}, \textcolor{denim}{$\mathbf{V}_4$}, \textcolor{denim}{$\mathbf{V}_5$}, and \textcolor{denim}{$\mathbf{V}_6$}), and a fixed textual description \texttt{``Forged Face''}.
    LF tokens are introduced in the CLIP text encoder and detailed in Fig.~\ref{fig_archi}\textcolor{red}{b} and Sec.~\ref{subsec:forgery_prompt_learn}.
    (b) The Bridge Adapter connects the CLIP image encoder to the deepfake detector. It integrates with an LLM and a classification head, which output textual explanations and a predicted fake score, respectively. 
    }
    \label{fig_overall_2}
\end{figure*}

Recently,~the~powerful~capability of vision-language models, \textit{e.g.,}~CLIP~\cite{radford2021learning}, also inspired efforts in detecting deepfakes.
For example, DDVQA-BLIP~\cite{zhang2024common} reformulates deepfake detection as an explanation generation task using a vision-language model~\cite{li2022blip}, which enhances interpretability through natural language descriptions (Fig.~\ref{fig:overall}\textcolor{red}{b}). 
In addition, several binary detectors~\cite{cozzolino2024raising, ojha2023towards, sha2023fake} leverage CLIP’s robust recognition capabilities to achieve impressive performance.
However, three key limitations remain in these works.
First, DDVQA-BLIP relies on a general text-generation model without dedicated mechanisms for deepfake detection, resulting in lower detection accuracy 
compared to conventional binary detectors.
Secondly, prior CLIP-based detectors often lack effective input text prompts to describe diverse forgeries, restricting the adaptation of CLIP's multi-modal learning ability in the detection task.
Third, while CLIP’s open-set recognition capability --- enabling it to identify diverse visual semantics --- is successfully combined with LLMs in domains like document parsing~\cite{ye2023mplug, liu2024textmonkey, hu2024mplug} and medical diagnosis~\cite{li2024llava, moor2023med, zhang2023pmc}, its integration with LLMs for deepfake detection remains largely unexplored.

To address these limitations, we propose a multi-modal face forgery detector (M2F2-Det), which contains dedicated forgery detection mechanisms for accurate detection and generating convincing textual explanations (Fig.~\ref{fig:overall}\textcolor{red}{c}): explanations enhance detection's trustworthiness, while accurate detection, in turn, supports reliable explanation generation through effective representation learning.
Moreover, the M2F2-Det introduces Forgery Prompt Learning, an efficient adaptation strategy that produces discriminative text embeddings for diverse forged face images.
We also introduce a Bridge Adapter to leverage the frozen CLIP image encoder, enhancing M2F2-Det’s detection performance and facilitating its integration with the LLM for textual explanation generation.

\begin{figure*}[t]
  \centering
    \begin{overpic}[width=1\linewidth]{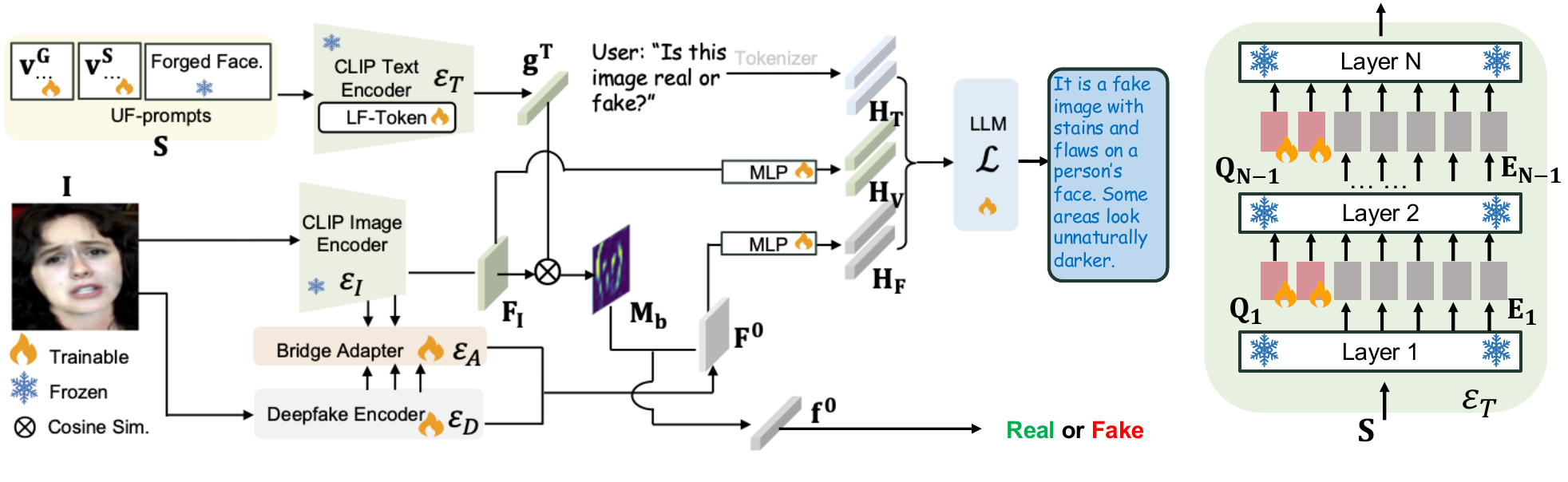}
    \put(34,0){(a)}
    \put(87,0){(b)}
    \end{overpic}
    \caption{    
    (a) The multi-modal face forgery detector (M2F2-Det) comprises
    pre-trained CLIP image and text encoders (\textit{i.e.}, $\mathcal{E}_{I}$ and $\mathcal{E}_{T}$), a deepfake encoder, as well as an LLM. 
    Given the universal forgery prompts (UF-prompts) as input, $\mathcal{E}_{T}$ generates a global text embedding, \textit{e.g.}, $\mathbf{g}^\text{T}$, that guides the generation of a forgery attention mask, \textit{e.g.}, $\mathbf{M}_b$. 
    The deepfake encoder utilizes the bridge adapter, \textit{i.e.}, $\mathcal{E}_A$, for detecting face forgeries (Sec.~\ref{subsec:forgery_loc_det}), while the LLM generates explanations conditioned on a frequency token $\mathbf{H}_\text{F}$ transformed from the forgery representation ($\mathbf{F}^0$) (Sec.~\ref{subsec:ff_it}).
    (b) In the CLIP text encoder, we introduce trainable layer-wise forgery tokens as inputs to each Transformer~\cite{vaswani2017attention} encoder layer.
    }
    \label{fig_archi}
\end{figure*}

Forgery Prompt Learning (FPL) comprises two key components: \textit{universal forgery prompts} (UF-prompts) and \textit{layer-wise forgery tokens} (LF-tokens) (Fig.~\ref{fig_overall_2}\textcolor{red}{a}).
First, UF-prompts include both general and specific forgery tokens:
general forgery tokens capture common forgery patterns and invariants shared across various manipulated facial images --- critical for generalizing to unseen forgeries; 
specific forgery tokens, by contrast, encode fine-grained, image-dependent artifacts, such as blurred eyes from attribute manipulation and blending boundaries from face swapping.
Secondly, we freeze the CLIP text encoder and introduce trainable layer-wise forgery tokens as inputs to its Transformer~\cite{vaswani2017attention} layers (Fig.~\ref{fig_archi}\textcolor{red}{b}). 
These task-specific tokens improve CLIP’s adaptability to deepfake while largely preserving the recognition ability of its pre-trained weights.

We further propose a Bridge Adapter (Bri-Ada) to harness capabilities of the pre-trained CLIP image encoder for both forgery detection and explanation generation. 
As depicted in Fig.~\ref{fig_overall_2}\textcolor{red}{b}, the Bri-Ada reuses intermediate features from the CLIP image encoder, preserving its foundational strengths in representation learning, which proves generalizable enough to identify unseen forgeries~\cite{cozzolino2024raising,ojha2023towards,sha2023fake}. 
To enhance domain-specific discrimination, Bri-Ada incorporates a deepfake encoder that provides forgery-aware knowledge, enabling the construction of more robust and effective visual representations for deepfake detection.
In addition, Bri-Ada is employed jointly with FPL in M2F2-Det (Fig.~\ref{fig_archi}\textcolor{red}{a}).
Such text embeddings generated by FPL are used to produce forgery attention maps, serving as prior knowledge to guide forgery identification.
Furthermore, Bri-Ada’s output is connected to the LLM, which leverages CLIP’s open-set recognition capability to translate visual features into textual explanations.
Specifically, Bri-Ada’s output is transformed into a frequency-based token.
This token then is concatenated with tokens from other modalities to guide the LLM in generating trustworthy explanations for deepfake detection, detailed in Sec.~\ref{subsec:ff_it}.
In summary, our contributions are:

$\diamond$ We propose a multi-modal face forgery detector, M2F2-Det, which innovatively outputs both deepfake detection scores and textual explanations, achieving remarkable detection accuracy and enhanced interpretability. 

$\diamond$ M2F2-Det introduces a Forgery Prompt Learning mechanism---automated and effective prompt learning tailored for deepfake detection---that transfers CLIP's powerful multi-modal learning ability into deepfake detection.

$\diamond$ M2F2-Det employs a Bridge Adapter that enhances the integration of LLM, facilitating the generation of trustworthy explanations for detection decisions.

$\diamond$ M2F2-Det achieves state-of-the-art (SoTA) deepfake detection performance, measured by $6$ datasets, showing the effectiveness of capturing diverse forgeries. 
It also obtains SoTA explanation generation performance on the DD-VQA dataset~\cite{zhang2024common}, both quantitatively and qualitatively.

\Section{Related Works}
\Paragraph{Deepfake Detection} 
The image forensics community develops various effective deepfake detection techniques, including data augmentation \cite{li2020face,liexposing,shiohara2022detecting,zhao2021learning,unified-detection-of-digital-and-physical-face-attacks}, frequency clues \cite{liu2021spatial,luo2021generalizing,wang2023dynamic}, disentanglement leanring~\cite{yan2023ucf,liang2022exploring,yang2019exposing_uadf,sun2024diffusionfake,sun2024continual,pan2024towards}, specified networks \cite{wang2021representative,tan2024data,jia2024can,RED_journal,tracing-hyperparameter-dependencies-for-model-parsing-via-learnable-graph-pooling-network}, and biometric information analysis~\cite{li2018ictu,sun2021improving,zhu2021face}. 
These works belong to the conventional deepfake detector that outputs binary prediction scores. 
Recently, DDVQA-BLIP~\cite{zhang2024common} defines a paradigm that generates textual explanations for deepfake detection, enhancing interpretability.
In contrast, the M2F2-Det outputs both a prediction score and textual explanations, using the latter to enhance detection interpretability, while the prediction score helps more convincing explanations.
Moreover, unlike prior CLIP-based forgery detectors~\cite{ojha2023towards, sha2023fake, cozzolino2024raising} that only rely on detection capabilities of pre-trained CLIP, our M2F2-Det further integrates 
the open-set visual recognition ability of the CLIP image encoder for enhanced interpretability.

\Paragraph{Prompt Learning} 
Prompt learning offers an efficient strategy adapting the pre-trained CLIP to downstream tasks~\cite{sun2022dualcoop, zhou2022conditional,kim2023zegot,jia2022visual,zhou2023anomalyclip,zhang2024tamm,xiufeng_lamma_detection}.~For example, CoOp~\cite{zhou2022learning} and CoCoOp~\cite{zhou2022conditional} integrate continuous prompts in the textual space, enhancing the pre-trained CLIP's generalizability.
Meanwhile, MaPLE~\cite{khattak2023maple} and VPT~\cite{jia2022visual} modify the learning procedure in visual spaces by altering the CLIP image encoder. 
These works learn global image information for recognition tasks while our FPL conducts the pixel-wise task to localize facial forgeries. 

\noindent\textbf{Multimodal Vision-Language Models}
(MLLMs)~\cite{yin2023survey,li2022blip,li2023blip,liu2024visual} use generative capabilities from LLMs~\cite{touvron2023llama,zhang2022opt} to obtain impressive performance across a wide range of tasks. 
For example, early studies concentrate on generating text-based content grounded on image, video, and audio~\cite{liu2024visual,li2023videochat,awadalla2023openflamingo,zhang2023video,deshmukh2023pengi,wu2023cheap}. 
Recently, MLLMs broaden applications to more complex downstream domains, including embodied AI~\cite{peng2023kosmos,zhang2024spartun3d}, document parsing~\cite{ye2023mplug,liu2024textmonkey,hu2024mplug}, and medical diagnosis~\cite{li2024llava,moor2023med,zhang2023pmc}. 
We propose a frequency token that implicitly aligns deepfake domain knowledge with MLLM, bridging the gap between language and subtle facial forgeries. 
\Section{Method}
\label{sec:method_overall}

\SubSection{Preliminaries}
\label{sec:preliminary}
We denote the input image and text prompts as $\mathbf{I}$ and $\mathbf{S}$, respectively. 
The proposed M2F2-Det utilizes CLIP’s image and text encoders, $\mathcal{E}_{I}$ and $\mathcal{E}_{T}$, together with a deepfake encoder, $\mathcal{E}_{D}$, for forgery detection.
Also, a large language model (\textit{i.e.}, $\mathcal{L}$) is employed to generate textual explanations. 

\Paragraph{Prompt Learning} Contrastive Language-Image Pre-training, known as CLIP~\cite{radford2021learning}, is a large-scale vision-language foundation model that has powerful zero-shot classification capabilities. 
Given a set of $K$ text prompts $\{\mathbf{S}_1, \mathbf{S}_2, \ldots, \mathbf{S}_K\}$, CLIP can estimate the likelihood that $\mathbf{I}$ corresponds to each of these text prompts:

\begin{equation}
p(y|\mathbf{I}) = \frac{\exp \left( \langle \mathcal{E}_{I}(\mathbf{I}), \mathcal{E}_{T}(\mathbf{S}_k) \rangle / \tau \right)}{\sum_{k=1}^{K} \exp \left( \langle \mathcal{E}_{I}(\mathbf{I}), \mathcal{E}_{T}(\mathbf{S}_k) \rangle / \tau \right)},
\label{eq:clip_form}
\end{equation}
where $\langle \cdot, \cdot \rangle$ and $\tau$ denote cosine similarity and a temperature hyper-parameter, respectively. 
To enhance the pre-trained CLIP's performance on downstream tasks, CoOp~\cite{zhou2022learning} proposes the prompt learning strategy that uses trainable tokens to automatically learn effective text prompts as follows:
\begin{equation}
    \mathbf{S}_k = [\mathbf{v}_1][\mathbf{v}_2] \ldots [\mathbf{v}_{n}][\texttt{class}_k],
\label{eq:trainable_token_form}
\end{equation}
where $[\mathbf{v}_1][\mathbf{v}_2] \ldots [\mathbf{v}_{n}]$ ($\mathbf{v}_{n} \in \mathbb{R}^{d}$) are trainable tokens, and $[\texttt{class}_k]$ represents the fixed and non-trainable class name of the $k$-th class.

\Paragraph{Visual Instruction Tuning} MLLMs tackle complicated reasoning tasks by generating responses based on visual and textual inputs. 
In general, a MLLM consists of three main components: 
1) a pre-trained image encoder, \textit{e.g.}, $\mathcal{E}_I$, that transforms $\mathbf{I}$ into a set of visual features. 
2) a projector, \textit{e.g.}, \texttt{MLP} layers, that converts visual features to visual tokens denoted as $\mathbf{H}_\text{V} \in \mathbb{R}^{N \times D}$.
3) an LLM, \textit{i.e.}, $\mathcal{L}$, that generates free-form responses in an auto-regressive manner when prompted with $\mathbf{H}_\text{V}$ and textual tokens $\mathbf{H}_\text{T} \in \mathbb{R}^{M \times D}$. 
$\mathbf{H}_\text{T}$ is generated by the tokenizer that takes user-input questions.
Let us define target answer tokens as $\mathbf{X}_\text{A} = [\mathbf{x}_1, \mathbf{x}_2,...,\mathbf{x}_\text{z}] \in \mathbb{R}^{Z\times D}$, where $Z$ represents the sequence length, then the probability of generating $\mathbf{X}_\text{A}$ becomes
\begin{equation}
p(\mathbf{X}_\text{A} \mid \mathbf{H}_\text{V}, \mathbf{H}_\text{T}) = \prod_{\text{z}=1}^{Z} p_\theta(\mathbf{x}_\text{z} \mid \mathbf{H}_\text{V}, \mathbf{H}_{\text{T},<\text{z}}, \mathbf{x}_{\text{A},<\text{z}}),
\label{eq:mllm_form}
\end{equation}
where $\theta$ are trainable parameters; $\mathbf{H}_{\text{T}, <\text{z}}$ and $\mathbf{x}_{\text{A}, <\text{z}}$ are instruction and answer tokens in all turns before the current prediction token $\mathbf{x}_\text{z}$, respectively.




\SubSection{Multi-modal Face Forgery Detector}
\label{sec:M2F2-det}

\SubSubSection{Forgery Prompt Learning}
\label{subsec:forgery_prompt_learn}
Forgery Prompt Learning captures forgeries via Universal Forgery Prompts \textit{e.g.}, UF-prompts, that contain two types of learnable tokens, \textit{e.g.}, general-forgery and specific-forgery tokens, \textit{e.g.}, $[\mathbf{v}^\text{G}]\in \mathbb{R}^{d}$ and $[\mathbf{v}^\text{S}]\in \mathbb{R}^{d}$, respectively.
Formally, we use \texttt{MLP} layers to transform the global visual embedding $\mathbf{g}^\text{I}\in\mathbb{R}^{d} = \mathcal{E}_I(\mathbf{I})$ into $[\mathbf{v}^\text{S}]$, which helps inject image-dependent information into $[\mathbf{v}^\text{S}]$. 
Consequently, $[\mathbf{v}^\text{G}]$ and $[\mathbf{v}^\text{S}]$ are optimized together to leverage the powerful multi-modal representation capability of the CLIP for capturing both general and image-specific forgery patterns.
Next, without loss of generality, we use \texttt{``forged face''} as the generic textual description for various input face images, which stabilizes the training empirically.
Therefore, we construct UF-prompts as:
\begin{equation}
    \mathbf{S} = [\mathbf{v}^\text{G}_1]\ldots[\mathbf{v}^\text{G}_m][\mathbf{v}^\text{S}_1]\ldots[\mathbf{v}^\text{S}_u]\texttt{[forged][face]},
\end{equation}
where $m\in\{0,1...M\}$, $u\in\{0,1...U\}$;~\texttt{forged} and \texttt{face} are non-trainable tokens converted by fixed words.

To enhance the conversion of $\mathbf{S}$ into textual embeddings that facilitate CLIP's adaptation, similar to the prior work~\cite{jia2022visual}, we introduce trainable layer-wise forgery tokens as inputs to each Transformer encoder layer of $\mathcal{E}_T$ while keeping its pre-trained weights frozen, depicted in Fig.~\ref{fig_archi}\textcolor{red}{b}.
More formally, for $\mathcal{E}_T$'s $(r+1)$-th layer, \textit{e.g.}, $\mathcal{E}^{r+1}_T$, we denote collections of input $O$ layer-wise forgery tokens and $P$ ordinary tokens as
$\mathbf{Q}_r = [\mathbf{q}_1, \mathbf{q}_2,...,\mathbf{q}_o] \in \mathbb{R}^{O\times d}$ and $\mathbf{E}_r = [\mathbf{e}_1, \mathbf{e}_2,...,\mathbf{e}_p] \in \mathbb{R}^{P\times d}$, respectively.
Then, $\mathcal{E}_T$'s forward propagation of taking $\mathbf{S}$ is:
\begin{align}
    \mathbf{E}_0 &= \texttt{Embed}(\mathbf{S}),\\
    \left[ \_, \mathbf{E}_{r+1} \right] &= \mathcal{E}^{r+1}_{T}\left( \mathbf{Q}_{i}, \mathbf{E}_{i} \right),
\end{align}
where $r$ is the layer index and \texttt{Embed} denotes the procedure that converts $\mathbf{S}$ into the $d$-dimensional latent space with positional embeddings. 

Specifically, taking $\mathbf{S}$ as the input, the $\mathcal{E}_{T}$ outputs the global textual embedding, denoted as $\mathbf{g}^\text{T} \in \mathbb{R}^{d}$. 
Meanwhile, we obtain output features from the last layer of $\mathcal{E}_{I}$, 
denoted as $\mathbf{F}_\text{I}\in\mathbb{R}^{N\times d} = \mathbb{R}^{W\times H \times d}$.
Then, a forged attention map, \textit{i.e.}, $\mathbf{M}_{b} \in \mathbb{R}^{W\times H}$, can be obtained by calculating the text-to-image score for its each patch, \textit{i.e.},
$\mathbf{M}^{ij}_{b} = \langle \mathbf{F}^{ij}_\text{I}, \mathcal{E}_{T}(\mathbf{S}) \rangle$, 
where $\mathbf{M}^{ij}_{b}$ and $\langle \cdot, \cdot \rangle$  represent text-to-image score for \( (i, j) \)-th patch of $\mathbf{M}_{b}$ and a cosine similarity operation, respectively.
Next, we use $\mathbf{M}_{b}$ to help detect global image-level forgeries because it provides prior knowledge of the spatial forgery location, and this technique is similar to prior works~\cite{zhao2021multi,hifi_net_xiaoguo,stehouwer2019detection}.
Please note unlike prior works~\cite{zhou2022learning,zhou2022conditional,kim2023zegot,jia2022visual} that adopt the prompt learning to capture global information of images, we innovatively propose the FPL to conduct a pixel-wise task of localizing forged regions (\textit{e.g.}, $\mathbf{M}_{b}$).
\begin{figure}[t]
    \centering
    \includegraphics[width=1.\linewidth]{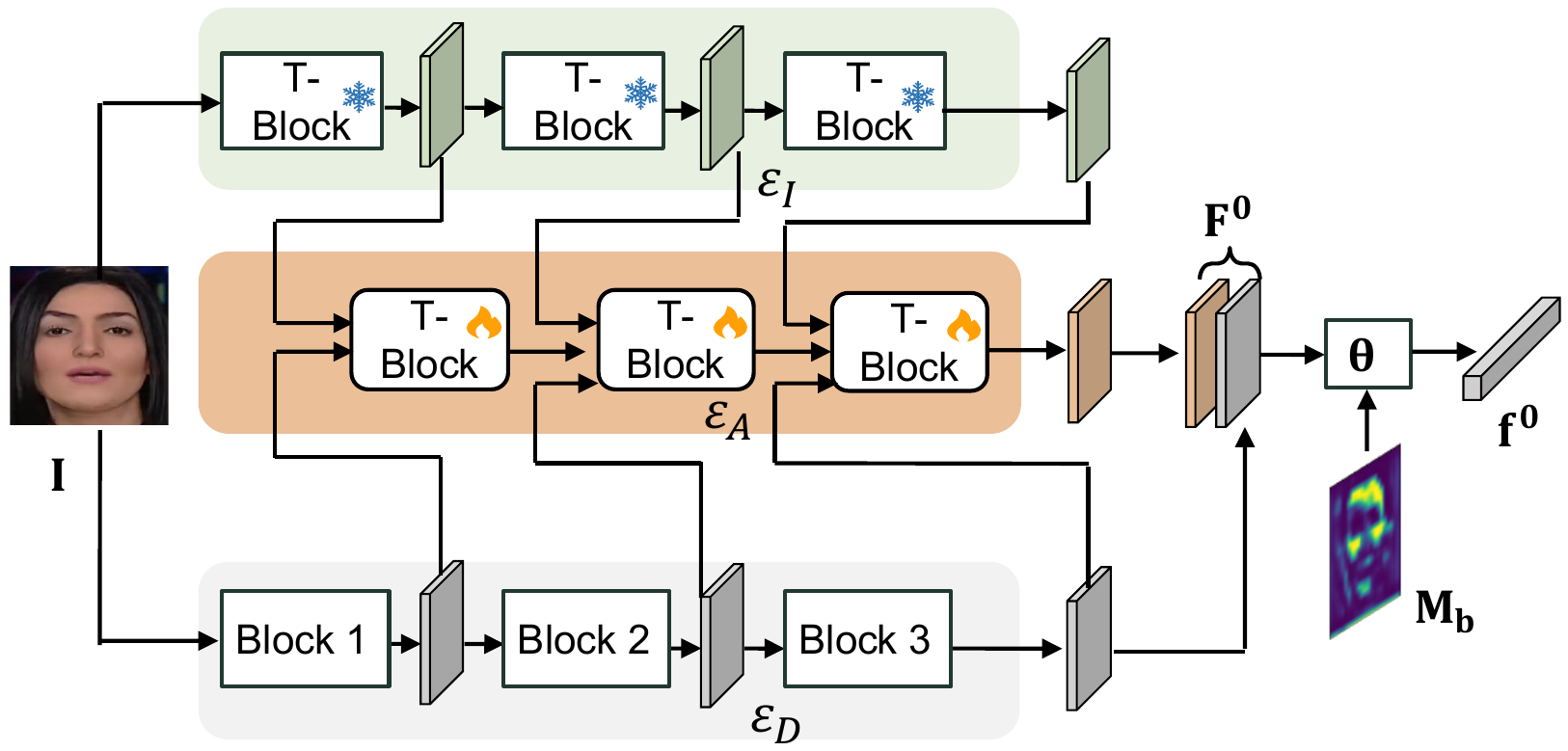}
    \caption{The illustration on the Bridge Adapter, in which $\Theta$ represents the transformation conducted by Eq.~\ref{eq:conv_pool}. [Key: T-Block: transformer encoder block; Block: convolution block.] \vspace{-3mm}}
    \label{fig:b_adapter}
\end{figure}

\SubSubSection{Bridge Adapter}
\label{subsec:forgery_loc_det}
As depicted in Fig.~\ref{fig:b_adapter}, the Bridge Adapter (Bri-Ada), \textit{i.e.}, $\mathcal{E}_{A}$, is composed of Transformer encoder blocks and takes intermediate features from both $\mathcal{E}_{I}$ and $\mathcal{E}_{D}$ as inputs.
In this way, $\mathcal{E}_{A}$ fully leverages the detection and open-set recognition capabilities of $\mathcal{E}_{I}$, further enriched by domain knowledge from $\mathcal{E}_{D}$ to produce more robust and effective representations for deepfake detection.

Bri-Ada is jointly used with FPL, which generates local forged attention maps.
Specifically, we concatenate feature maps output from $\mathcal{E}_{D}$ and $\mathcal{E}_{A}$ 
into a fused feature map $\mathbf{F}^{0}\in \mathbb{R}^{w\times h\times c}$, as illustrated in Fig.~\ref{fig:b_adapter}.
We then use $\mathbf{M}_{b}$ and $\mathbf{F}^{0}$ to obtain refined forgery vector, \textit{e.g.}, ${\mathbf{f}^{0}} \in \mathbb{R}^{d}$, as the final forgery representation for deepfake detection.
Specifically, we have 
\begin{equation}
    \mathbf{f}^{0} = \texttt{AVGPOOL}(\texttt{CONV}(\mathbf{F}^{0} \odot \mathbf{M}_{b})),
    \label{eq:conv_pool}
\end{equation}
where \texttt{AVGPOOL} and \texttt{CONV} represent the average pooling operation and convolution layers, respectively. 

The joint use of FPL and $\mathcal{E}_A$ for detection has two key advantages.
First, FPL and $\mathcal{E}_A$ mutually benefit each other. 
For global detection, $\mathcal{E}_A$ reuses intermediate features from $\mathcal{E}_D$ and then employs $\mathbf{M}_{b}$ as prior knowledge of forgery regions.
This encourages FPL to update $\mathbf{S}$ and $\mathbf{Q}_{i}$, such that $\mathbf{g}^\text{T}$ can be used to generate accurate forged attention maps.
For generating forgery attention maps, FPL needs feedback from $\mathcal{E}_{A}$ and $\mathcal{E}_{D}$, on if its generated $\mathbf{M}_{b}$ enhances binary detection.
Secondly, M2F2-Det is only supervised by the binary ground truth label, indicating that $\mathbf{M}_{b}$ is learned via an efficient unsupervised manner.

\SubSubSection{Forgery Explanation Module}
\label{subsec:ff_it}
The Forgery Explanation Module helps the M2F2-Det generate texts, as depicted in Fig.~\ref{fig_archi}\textcolor{red}{a}.
These generated texts contain the judgment and explanation, which claims if the image is forged and explicitly describes the rationale behind this decision, respectively.
Specifically, we obtain the representation $\mathbf{F}^{0}$ for the detection task.
We convert it into $\mathbf{H}_\text{F} \in \mathbb{R}^{N\times D}$.
As a result, $\mathbf{H}_\text{F}$ informs the LLM (\textit{e.g.}, $\mathcal{L}$) if the input image is fake.
Meanwhile, we transform $\mathcal{E}_{I}$'s output feature into visual tokens $\mathbf{H}_\text{V}$, which helps $\mathcal{L}$ describe the facial pattern.
Both $\mathbf{H}_\text{F}$ and $\mathbf{H}_\text{V}$ are fed into $\mathcal{L}$ for explanation generation. 
Therefore, we update the Eq.~\ref{eq:mllm_form} into as follows:
\begin{equation}
    p(\mathbf{X}_\text{A} \mid \mathbf{H}_\text{V}, \mathbf{H}_\text{F}, \mathbf{H}_\text{T}) = \prod_{\text{z}=1}^{Z} p_\theta(\mathbf{x}_\text{z} \mid \mathbf{H}_\text{V}, \mathbf{H}_\text{F}, \mathbf{H}_{\text{T},<\text{z}}, \mathbf{x}_{\text{A},<\text{z}}).
    \label{eq:trainable_freq}
\end{equation}



\SubSection{Train and Inference}
\label{sec:train_infer}

\Paragraph{Training} First, we train the deepfake encoder $\mathcal{E}_D$ as well as  $\mathbf{S}$ and $\mathbf{Q}$ in FPL, such that M2F2-Det can perform the binary deepfake classification. 
We minimize the cross entropy distance between binary classification probability $p(y|\mathbf{I})$ and a ground truth categorical $\hat{y}$.

Secondly, we align $\mathbf{H}_\text{V}$ and $\mathbf{H}_\text{F}$ with the input space of a frozen LLM, such that outputs from $\mathcal{E}_I$ and $\mathcal{E}_D$ can be interpreted by the LLM. 
More formally, we maximize the likelihood defined in Eq.~\ref{eq:trainable_freq} via only training MLP layers while freezing other components
in this stage.

Thirdly, to better tame the LLM for explanation generation, we again keep the entire model frozen while only updating \texttt{MLP} layers and LLM based on Eq.~\ref{eq:trainable_freq}.  
Trainable parameters, \textit{e.g.}, $\theta$, thus become \texttt{MLP} layers and a subset of LLM's parameters.
Please note we use LoRA for efficient LLM fine-tuning.

In second and third-stage training, we use the DD-VQA~\cite{zhang2024common} dataset that contains high-quality image-text pairs annotated by the Amazon mechanical Terk.
The DD-VQA dataset consists of $14,782$ question-answer pairs using the train/test IDs from FF++~\cite{rossler2019faceforensics}.
This results in $13,559$ question-answer pairs for training and $1,223$ pairs for testing. 
Note that the second and third training stages are similar to LLaVA~\cite{liu2024visual}, and the difference is we align one more representation, \textit{i.e.}, $\mathbf{H}_\text{F}$. 

\Paragraph{Inference} Given the input image and user-input questions, we produce the binary result and textual explanations.
User-input questions can be flexible, such as \texttt{``Determine the authenticity of the image.''} and \texttt{``Is this image real or fake?''}. 
\begin{table}[t]
\centering
    \scalebox{1}{
        \begin{tabular}{l|c|c}
        \toprule
        \textbf{Dataset} & \textbf{Real Samples} & \textbf{Fake Samples}\\ \hline
        FF++~\cite{rossler2019faceforensics} & $1,000$ V & $4,000$ V\\ \hline
        CDF~\cite{li2019celeb}  & $590$ V & $5,639$ V\\ \hline
        DFD~\cite{deepfakedetection} & $363$ V & $3,068$ V \\ \hline
        WDF~\cite{zi2020wilddeepfake}  & $3,805$ I & $3,509$ I \\ \hline
        DFDC~\cite{google_dfd} & $1,131$ V & $4,113$ V\\ \hline
        FFIW~\cite{zhou2021face} & $10,000$ V & $10,000$ V \\ \bottomrule
        \end{tabular} 
    }
\caption{Six datasets used for evaluating detection performance. [Key: V: Video; I: Image].}
\label{tab:datasets_overview}
\end{table}

\begin{table*}[t]
\centering
    \scalebox{0.9}{
    \begin{tabular}{c| c| cc| cc| cc| cc}
        \hline
        Methods & Venue & \multicolumn{2}{c|}{FF++ (c23)} & \multicolumn{2}{c|}{FF++ (c40)} & \multicolumn{2}{c|}{Celeb-DF} & \multicolumn{2}{c}{WDF} \\
        \cline{3-10}
        &&\multicolumn{8}{c}{\textit{Metric:} Acc (\% $\uparrow$) / AUC (\% $\uparrow$)}\\
        \hline
        RFM~\cite{wang2021representative} &CVPR21& $95.69$ & $98.79$ & $87.06$ & $89.83$ & $97.96$ & $99.94$ & $77.38$ & $83.92$ \\
        \rowcolor{gray!30}Freq-SCL~\cite{li2021frequency} &CVPR21& $96.69$ & $99.28$ & $89.00$ & $92.39$ & – & – & – & – \\
        Add-Net~\cite{zi2020wilddeepfake} &ACMMM20& $96.78$ & $97.74$ & $87.50$ & $91.01$ & $96.93$ & $99.55$ & $76.25$ & $86.17$ \\
        \rowcolor{gray!30}F3-Net~\cite{qian2020thinking} &ECCV20& $97.52$ & $98.10$ & $90.43$ & $93.30$ & $95.95$ & $98.93$ & $80.66$ & $87.53$ \\
        MultiAtt~\cite{zhao2021multi} &CVPR21& $97.60$ & $99.29$ & $88.69$ & $90.40$ & $97.92$ & $\color{red}\mathbf{99.94}$ & $82.86$ & $90.71$ \\
        \rowcolor{gray!30}RECCE~\cite{cao2022end} & CVPR22& $97.06$ & $99.32$ & $91.03$ &$\color{blue}{95.02}$& $\color{blue}{98.59}$ & $\color{red}\mathbf{99.94}$ & $\color{blue}83.25$ & $\color{blue}92.02$ \\
        TALL~\cite{xu2023tall} & ICCV23& $\color{blue}{98.65}$ & $\color{red}\mathbf{99.87}$ & $\color{blue}{92.82}$ & $94.57$ & $97.57$ & $98.55$ & - & - \\
        \rowcolor{gray!30}DDVQA-BLIP~\cite{zhang2024common} & ECCV24 & $80.69$ & -  & $72.73$ & -  & - & - & - & - \\
        \hline
       M2F2-Det & & $\color{red}\mathbf{98.79}$ & $\color{blue}{99.34}$ & $\color{red}\mathbf{93.83}$ & $\color{red}\mathbf{96.58}$ & $\color{red}\mathbf{98.98}$ & $\color{blue}{99.92}$ & $\color{red}\mathbf{86.05}$ & $\color{red}\mathbf{93.14}$ \\
        \hline
    \end{tabular}}
\vspace{-0.5mm}
\caption{Intra-dataset Detection Performance. Results of prior works are mainly cited from \cite{cao2022end,xu2023tall}. 
[Key: \textbf{\color{red}{Best}}, \color{blue}{Second Best}].}
\label{tab:intra-dataset-protocol}
\end{table*}

\begin{table*}[t]
\centering
\scalebox{0.9}{
\begin{tabular}{c|c|cc|cccc}
\hline
\multirow{2}{*}{Method} & \multirow{2}{*}{Venue} & \multicolumn{2}{c|}{Training set} & \multicolumn{4}{c}{Test set AUC (\% $\uparrow$)} \\
& & Real & Fake & DFDC & FFIW & Celeb-DF & DFD  \\
\hline
LocalRL~\cite{chen2021local} & AAAI21 & $\checkmark$ & $\checkmark$  & $76.53$ & - & $78.26$ & $89.24$\\
\rowcolor{gray!30}CADDM~\cite{yan2023ucf} & CVPR23 & $\checkmark$ & $\checkmark$  & - & - & $93.88$ & $99.03$\\        
UCF~\cite{yan2023ucf} & ICCV23 & $\checkmark$ & $\checkmark$  & $80.50$ & -& $82.40$ & $94.50$ \\
\hline
\rowcolor{gray!30}SBI~\cite{shiohara2022detecting} & CVPR22 & $\checkmark$ & & $86.15$ & $84.83$ & $93.18$ & $97.56$ \\
AUNet~\cite{bai2023aunet} & CVPR23 & $\checkmark$ & & $86.16$ & $81.45$ & $92.77$ & $\color{red}\mathbf{99.22}$ \\
\rowcolor{gray!30}Seeable~\cite{larue2023seeable} & ICCV23 & $\checkmark$ & - & $86.30$ & - & $87.30$ & - \\
LAA-Net (w/ SBI)~\cite{nguyen2024laa} & CVPR24 & $\checkmark$ & & $86.94$ & - & $\color{red}{\mathbf{95.40}}$ & $\color{blue}{98.43}$ \\
\rowcolor{gray!30}FreqBlender~\cite{li2024freqblender} & NeurIPS24 & $\checkmark$ & & $\color{blue}{87.56}$ & $\color{blue}{86.14}$ & $94.59$ & - \\
\hline
M2F2-Det (w/SBI) & & $\checkmark$ & & $\color{red}{\mathbf{87.80}}$ & $\color{red}{\mathbf{88.70}}$ & $\color{blue}{95.10}$ & $97.70$ \\
\hline
\end{tabular}}%
\caption{Inter-dataset Detection Performance. Results of prior works are mainly
cited from \cite{shiohara2022detecting,nguyen2024laa,bai2023aunet,li2024freqblender}. [Key: \textbf{\color{red}{Best}}, \color{blue}{Second Best}]. \vspace{-2mm}}
\label{tab:inter-dataset-protocol}
\end{table*}
\Section{Experiment}
\label{sec:exp}

\SubSection{Setup} 
\label{exp:setup}

\Paragraph{Datasets}
We evaluate our method on both detection and explanation generation tasks.
For detection, we compare against existing detection approaches using datasets listed in Tab.~\ref{tab:datasets_overview}, including FaceForensics++ (FF++)~\cite{rossler2019faceforensics}, CelebDF~\cite{li2019celeb}, WildDeepfake (WDF)~\cite{zi2020wilddeepfake}, DFD~\cite{google_dfd}, DFDC~\cite{dolhansky2020deepfake}, and FFIW~\cite{zhou2021face}.
For explanation generation, we evaluate on the publicly available DD-VQA dataset~\cite{zhang2024common}.

\Paragraph{Metrics} 
First, we use Area Under the Curve (AUC) and accuracy to measure the detection performance. 
Second, for explanation generation performance, we evaluate \textit{judgment performance} and \textit{explanation quality}. 
For judgment performance, we extract keywords, \textit{e.g.}, $\texttt{"Fake"}$ and $\texttt{"Real"}$, from generated texts to compute accuracy and F1-Score.
To assess explanation quality, we employ standard natural language generation metrics such as BLUE-4~\cite{papineni2002bleu}, CIDEr~\cite{vedantam2015cider}, ROUGE\_L~\cite{rouge2004package}, METEOR~\cite{denkowski2014meteor}, and SPICE~\cite{anderson2016spice}. 
These metrics evaluate the similarity between generated and annotated textual answers comprehensively.
Additional evaluation details are provided in the supplementary material.

\Paragraph{Implmentation Details} 
We employ EfficientNet-B4~~\cite{tan2019efficientnet} as the deepfake detector, \textit{i.e.}, $\mathcal{E}_D$.
Additionally, we use the CLIP/ViT-L-patch14-336 model~\cite{dosovitskiy2020image} for the pre-trained CLIP image encoder ($\mathcal{E}_I$) and text encoder ($\mathcal{E}_T$).
The LLM is Vicuna-7b~\cite{chiang2023vicuna}, and more details are in the supplementary.


\begin{figure*}[t]
    \centering
    \begin{subfigure}[b]{0.35\textwidth}
        \centering
        \begin{tabular}{cccc}
            \hline
            \multirow{2}{*}{Method} & \multirow{2}{*}{\shortstack[c]{Fine-\\tuned}} & \multicolumn{2}{c}{Detection}\\
            & & Acc $\uparrow$ & F1 $\uparrow$ \\
            \hline
            DDVQA-BLIP & \cmark & $87.49$ & $90.07$ \\
           \rowcolor{gray!30}LLaVA & \xmark & $51.41$ & $37.04$ \\
           LLaVA & \cmark & $86.41$ & $92.10$\\
            \rowcolor{gray!30}Qwen & \xmark & $59.41$ & $51.04$\\
            InternVL & \xmark & $42.33$ & $47.35$ \\
            \hline
            \rowcolor{gray!30}M2F2-Det \textit{w/o} $\mathbf{H}_\text{F}$ & \cmark & $85.11$ & $84.21$ \\
            M2F2-Det & \cmark & $\mathbf{95.23}$ & $\mathbf{96.61}$ \\
            \hline
        \end{tabular}
        \caption{}
        \label{tab:sentence-generation-judgment}
    \end{subfigure}
    \hspace{1.2cm}
    \begin{subfigure}[b]{0.25\textwidth}
        \centering
        \includegraphics[width=\textwidth,height=4cm]{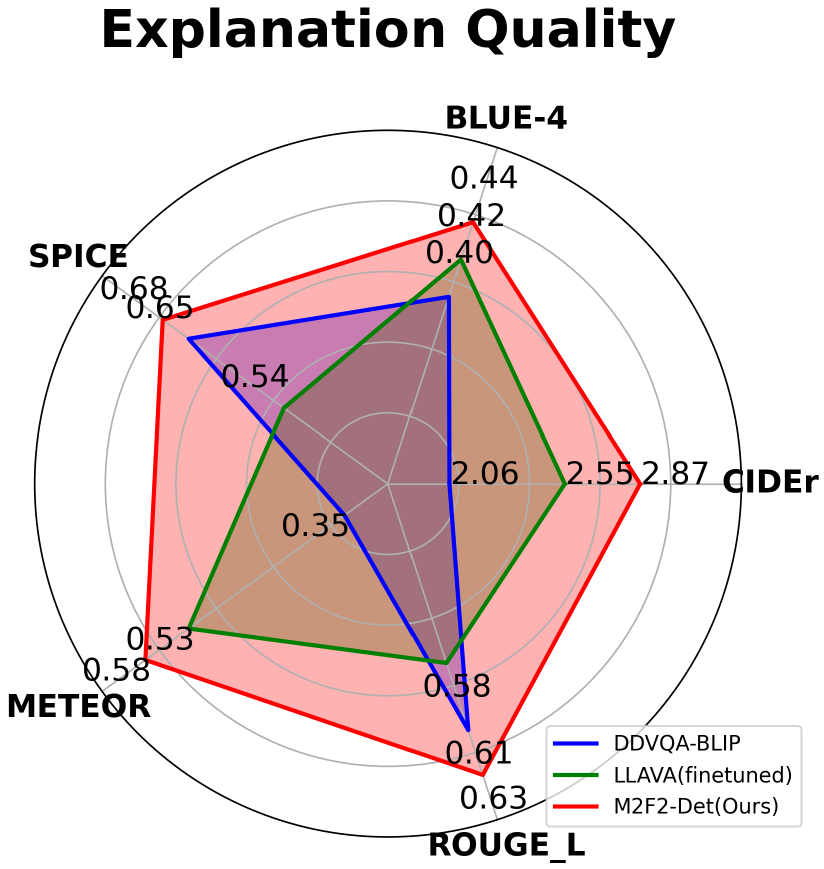} 
        \caption{}
        \label{tab:sentence-generation-quality}
    \end{subfigure}
    \hfill 
    \begin{subfigure}[b]{0.3\textwidth}
        \centering
        \includegraphics[width=1\textwidth]{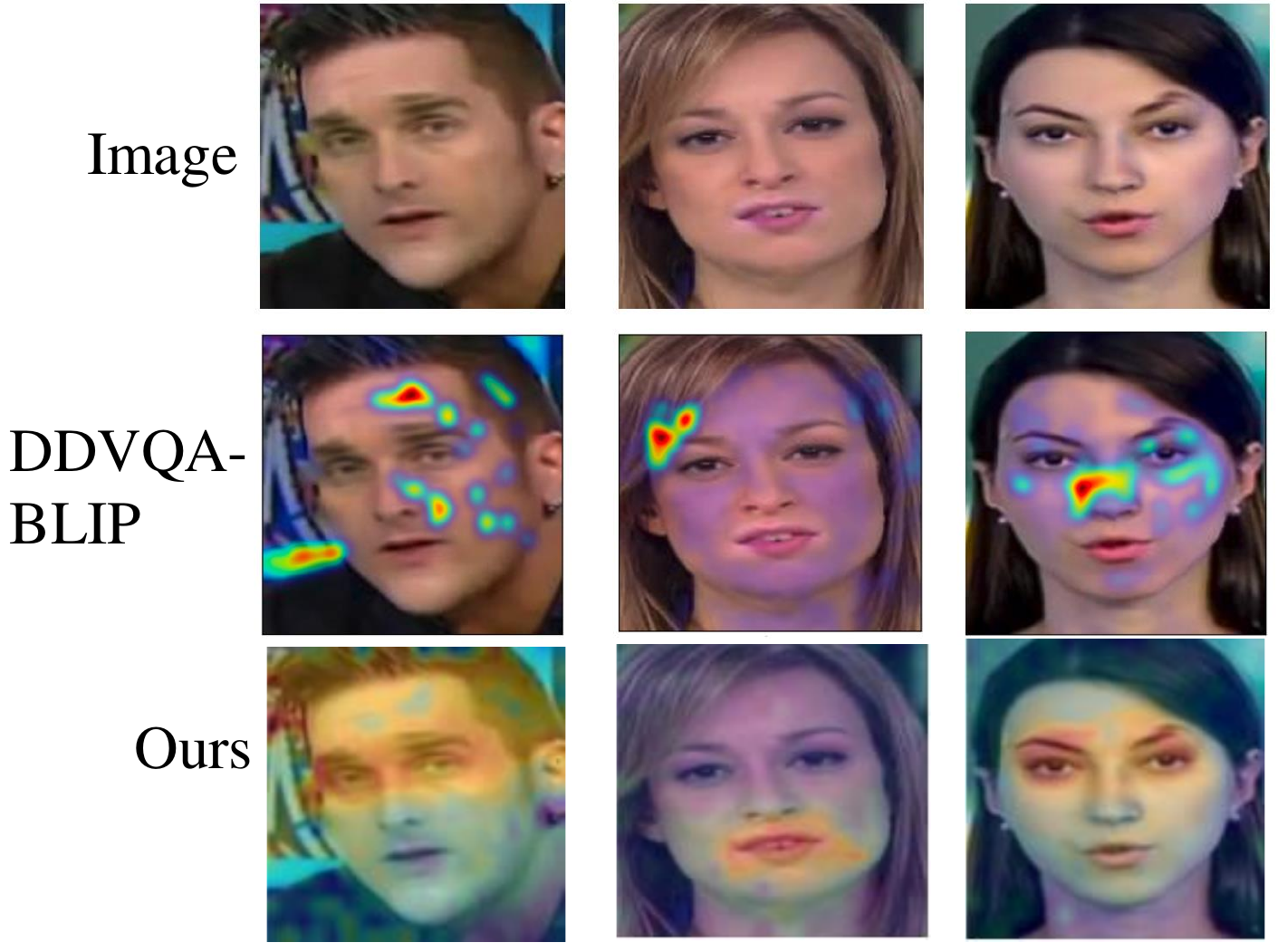} 
        \caption{}
        \label{tab:sentence-generation-heatmap}
    \end{subfigure}
    \vspace{-2mm}
    \caption{Explanation generation performance on DD-VQA. (a) Judgment performance. [Key: \textbf{Best results}, Acc: Accuracy, F1: F1 Score] (b) Explanation quality is measured by $5$ metrics. (c) Visualizations of forged attentions. 
    }
    \label{tab:sentence-generation}
\end{figure*}

\begin{figure*}[t]
    \centering
    \begin{overpic}[width=0.96\textwidth]{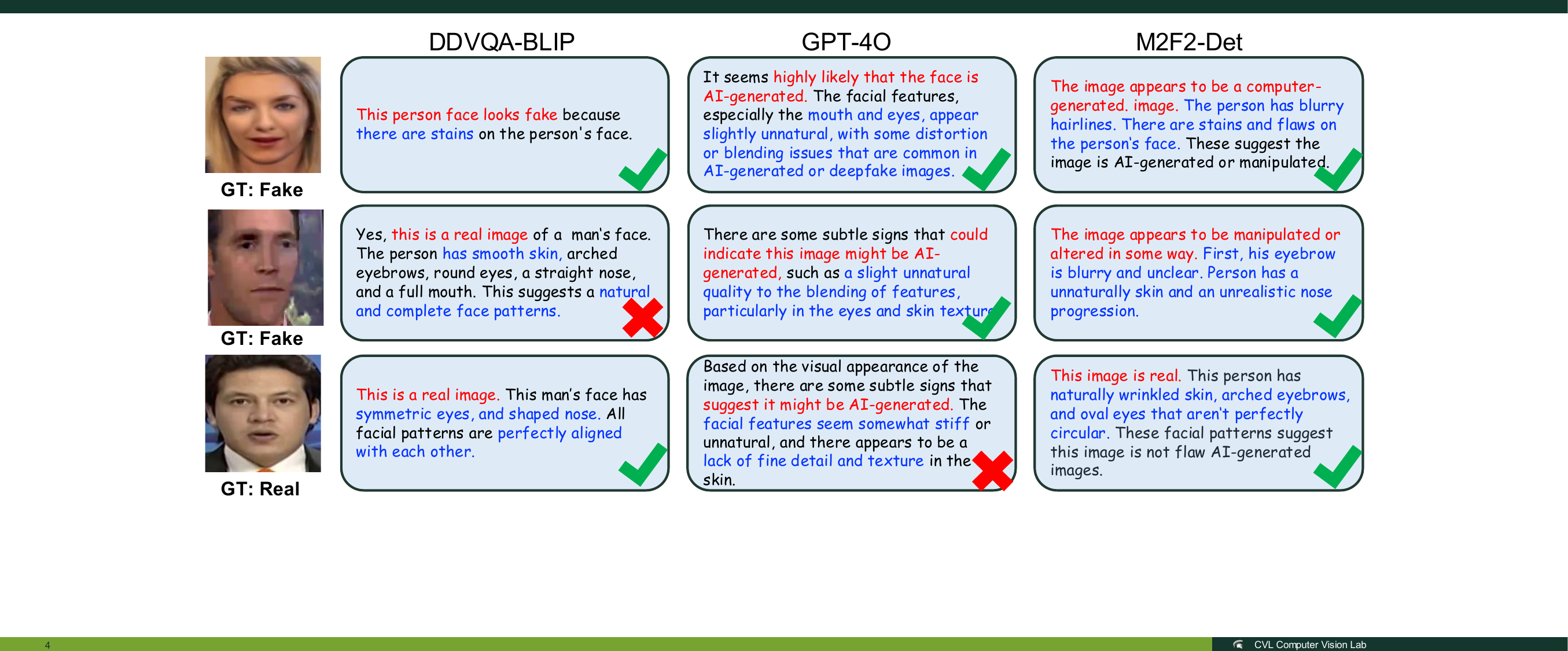}
    \put(-3,32){(a)}
    \put(-3,20){(b)}
    \put(-3,8){(c)}
    \end{overpic}
    \caption{Qualitative results of explanation generation from DDVQA-BLIP, GPT-4o, and M2F2-Det.\vspace{-2mm}
    }
    \label{fig:gen_sentence}
\end{figure*}
\SubSection{Detection Performance} 
\label{ex:detection}

Tab.~\ref{tab:intra-dataset-protocol} and \ref{tab:inter-dataset-protocol} report detection performance on intra- and inter-dataset setups, respectively.

\Paragraph{Intra-dataset performance}
Tab.~\ref{tab:intra-dataset-protocol} shows our M2F2-Det achieves the best overall detection performance in FF++. 
For example, in FF++ (c$40$), M2F2-Det has $1.01\%$ higher accuracy and $2.01\%$ higher AUC than the second-best method, \textit{i.e.}, TALL.
Please note that TALL takes multiple frames as inputs, which generally contain more forgery information than using one frame as the input. 
While being a detector that takes single-frame input, our method still outperforms TALL, 
indicating the effectiveness of the proposed Forgery Prompt Learning and Bridge Adapter.
Such effectiveness can be further demonstrated in Celeb-DF, in which our method surpasses TALL by $1.41\%$ accuracy and $1.37\%$ AUC score.

Furthermore, 
WDF collects diverse real-life forged faces from the web
on which our method outperforms the second-best performer, \textit{e.g.}, RECCE, by $2.80\%$ and $1.12\%$ in accuracy and AUC score, respectively.
One key difference between RECCE and our method is that we adapt the pre-trained CLIP image encoder, which is also trained on web samples. 
Such an adaptation improves M2F2-Det's detection ability by leveraging CLIP's robust representation, which proves generalizable enough to detect unseen forgeries~\cite{ojha2023towards,cozzolino2024raising}.
DDVQA-BLIP~\cite{zhang2024common} predicts fake images based on if keywords like ``\texttt{fake}'' in generated sentences, whereas its detection accuracy is $18.1\%$ and $21.1\%$ lower than us on FF++ c$23$ and c$40$, respectively.
This shows accurate deepfake detection requires a specific mechanism that learns forgeries, like using forged attention masks as local forged contexts, instead of applying the text-generation model~\cite{li2022blip}.

\begin{figure*}[t]
    \footnotesize
    \centering
    \includegraphics[width=1\linewidth]{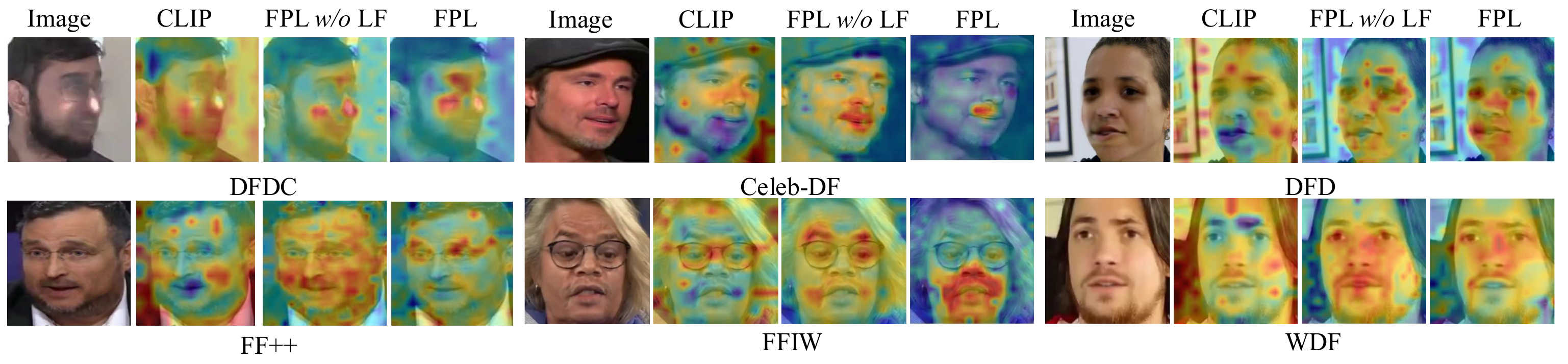}
    \vspace{-2mm}
    \caption{Forged attention maps on samples from $6$ datasets introduced in Tab.~\ref{tab:datasets_overview}. [Key: LF: layer-wise forgery tokens] \vspace{-1mm}}
    \label{fig:forged_map}
\end{figure*}

\Paragraph{Inter-dataset performance}~Tab.~\ref{tab:inter-dataset-protocol}~reports the performance, in which we follow prior works~\cite{shiohara2022detecting,nguyen2024laa} to train methods on real and pseudo-fake images from FF++.
Our M2F2-Det achieves $0.24\%$ and $2.56\%$ higher AUC than FreqBlender~\cite{li2024freqblender} on DFDC and FFIW, respectively.
We believe our method's superior generalization ability comes from the usage of a pre-trained CLIP image encoder.
Specifically, the CLIP is trained on diverse real-world web samples instead of a forgery detection dataset, making its learned features more generalizable and less overfitting on specific forgery patterns~\cite{cozzolino2024raising,ojha2023towards,sha2023fake}. 
Moreover, M2F2-Det outperforms AUNet with higher AUC scores on three datasets but performs worse on the DFD dataset.
AUNet learns forgery clues from relations between different facial action units.
Similarly, specific forgery information from action units could also be considered in M2F2-Det's FPL, which uses UF-prompts and layer-wise forgery tokens to adapt CLIP in discerning both general and specific facial forgeries. 
The learned text embedding is used to generate forged attention maps, as depicted in Fig.~\ref{fig:forged_map}, that act as local forged contexts to help the global detection task.

\SubSection{Explanation Generation Performance} 
\label{ex:explain} 
\Paragraph{Quantative Result} 
Fig.~\ref{tab:sentence-generation-judgment} reports explanation generation performance regarding \textit{judgment performance} and \textit{explanation quality}. 
First, from the judgment perspective, our method achieves the best accuracy and F1 score, which have $7.74\%$ higher accuracy than DDVQA-BLIP and $4.51\%$ higher F1 score than fine-tuned LLaVa.
This demonstrates the advantage of the proposed frequency-based token (\textit{i.e.}, $\mathbf{H}_\text{F}$), which helps our method generate correct descriptions based on learned deepfake domain knowledge.
The usage of $\mathbf{H}_\text{F}$ is different from previous MLLMs that only judge if visual artifacts exist in the color domain, which is less effective in capturing the discrepancy between real and fake images at frequency domains.
Fig.~\ref{tab:sentence-generation}\textcolor{red}{a} shows that a removal of $\mathbf{H}_\text{F}$ decreases M2F2-Det's judgment performance by $10.12\%$ in accuracy.
Secondly, Fig.~\ref{tab:sentence-generation-quality} shows that M2F2-Det achieves the best explanation quality.
Specifically, M2F2-Det has the best CIDEr score, which measures semantic consistency between generated sentences and ground truth (GT): Fig.~\ref{fig:gen_sentence} shows other works with erroneous judgments and explanations, different from GT's semantics, hence causing low CIDEr.
Also, ROUGE$\textunderscore$L measures if generated explanations summarize GT's information with lexical variations, and our results include more diverse phrases, \textit{e.g.}, \texttt{naturally~wrinkled~skin} in Fig.~\ref{fig:gen_sentence}'s 3rd sample.
Fig.~\ref{tab:sentence-generation-heatmap} shows learned forged maps used in the M2F2-Det can better identify artifacts and forgeries than the cross-attention mechanism of DDVQA-BLIP.

\Paragraph{Qualitative Performance} Fig.~\ref{fig:gen_sentence} reports qualitative results, where our M2F2-Det generates explanations with accurate judgments and convincing explanations for both real and fake images.
First, for image (a), DDVQA-BLIP cannot offer detailed explanations,
whereas our model provides convincing reasonings, \textit{i.e.}, \texttt{blurry hairlines}.
For (b), 
M2F2-Det again provides correct judgment and sophisticated explanations, {\it e.g.}, clearly identifying unnatural patterns of eye regions and unnatural skin textures. 
We believe that such performance superiority of M2F2-Det comes from the effective deepfake representation (\textit{i.e.}, $\mathbf{H}_\text{F}$).
Lastly, we also show generated sentences on a real image (c), showing M2F2-Det's effectiveness in discerning genuine faces.

\SubSection{Ablative Study and Analysis} 
\label{ex:ablative_study}
\begin{table}[t]
\centering
\begin{tabular}{l|ccc|cc}
\toprule
&\multicolumn{2}{c}{\textbf{Forgery PL}} & \multirow{2}{*}{Bri-Ada} & \multicolumn{2}{c}{Test set AUC (\%)} \\  
&UF-P & LF & & FF++(c40) & Celeb-DF \\
\midrule
$1$&&&&$91.03$&$65.78$ \\ \hline
\rowcolor{gray!30}$2$&&$\checkmark$&&$92.57$&$67.37$\\ \hline
$3$&$\checkmark$&&&$92.66$&$66.08$ \\ \hline
\rowcolor{gray!30}$4$&$\checkmark$&$\checkmark$&&$93.65$&$68.68$ \\ \hline
$5$&&&$\checkmark$&$93.80$&$70.71$\\ \hline
\rowcolor{gray!30}$6$&$\checkmark$&&$\checkmark$&$94.20$&$71.08$ \\ \hline
$7$&$\checkmark$&$\checkmark$&$\checkmark$&$\mathbf{96.58}$&$\mathbf{74.82}$\\ 
\bottomrule
\end{tabular}
\vspace{-1mm}
\caption{Ablation Study. Each model is trained by FF++(c40) and tested on FF++(c40) and Celeb-DF. [Keys: Forgery PL: Forgery Prompt Learning; UF-P: Universal Forgery Prompts; LF: layer-wise forgery tokens; Bri-Ada: Bridge Adapter.] \vspace{-3mm}
}
\label{table_ablation}
\end{table}

\Paragraph{Forged Attention Map} 
Tab.~\ref{table_ablation}'s line $\#1$ represents the deepfake detection baseline performance, \textit{e.g.}, EfficientNet-B4~\cite{tan2019efficientnet}.
Line $\#2$ and $\#3$ show layer-wise forgery tokens (LF tokens) and UF-prompts improve performance --- \textit{e.g.}, $1.54\%$ and $1.63\%$ higher AUC scores on FF++, respectively.
This shows that LF tokens and UF-prompts are effective designs to learn deepfake domain knowledge. 
Furthermore, in line $\#4$, we employ both UF-prompts and LF tokens, which is the full version of FPL, further increasing the performance of line $\#2$ by $1.08\%$ AUC on FF++.
This is because accurate forged attention maps can assist detection.
For example, in Fig.~\ref{fig:forged_map}'s first sample, when using FPL, the person's nose and eye areas are precisely identified as forged regions. 
Such visualizations further demonstrate the quality of forged attention maps obtained from the FPL.

\Paragraph{Bridge Adapter} The comparison between lines $\#1$ and $\#5$ demonstrates the effectiveness of the Bri-Ada --- increasing baseline's performance, \textit{e.g.}, line $1$, by $2.77\%$ AUC score on FF++. 
In addition, it enhances the generalizable detection ability and increases performance on Celeb-DF by $4.93\%$.
We believe this is because Bri. Ada. employs the CLIP image encoder that helps become less overfitting on
specific manipulation types of FF++ samples.
Line $\#6$ indicates the integration between Bri-Ada and FPL further increases detection performance.
Lastly, the full M2F2-Det (\textit{i.e.}, line $\#7$) achieves the best overall performance.

\Paragraph{Comparison to CLIP-based forgery detectors} Tab.~\ref{tab:subtable1} shows that M2F2-Det achieves $24.18\%$ and $19.74\%$ higher AUC scores than previous CLIP-based image forensic methods, Uni-Fake~\cite{ojha2023towards} and DEFAKE~\cite{sha2023fake}, respectively.
These two methods use the CLIP image encoder with simple architecture, \textit{e.g.}, linear layers and ResNet-18, which lack specific facial forgery mechanisms.
In contrast, the M2F2-Det not only employs the CLIP but also uses specified forgery detection mechanisms like FPL.

\Paragraph{Comparison to Prompt learning methods} We compute cosine similarities between CLIP image and text encoders' outputs for detection after applying different prompt learning schemes, and the performance is reported in Tab.~\ref{tab:subtable2}.
Our FPL achieves $9.31\%$ and $8.13\%$ higher AUC scores than CoOp~\cite{zhou2022learning} and CoCoOp~\cite{zhou2022conditional}, respectively.
This is because prior works are developed to adapt the CLIP to tasks that focus on recognizing semantics, which is different from deepfake detection.

\begin{table}[t]
    \centering
    \begin{subtable}[b]{0.45\linewidth}
        \centering
        \resizebox{1\textwidth}{!}{
        \begin{tabular}{cc}
            \hline
            Method & AUC \\
            \hline
            \rowcolor{gray!30}Uni.Fake~\cite{ojha2023towards} & $72.40$ \\
            DEFAKE~\cite{sha2023fake} & $76.84$ \\
            \rowcolor{gray!30}M2F2-Det & $\mathbf{96.58}$ \\
            \hline
        \end{tabular}}
        \subcaption{\vspace{-2mm}}
        \label{tab:subtable1}
    \end{subtable}
    \begin{subtable}[b]{0.45\linewidth}
        \centering
        \resizebox{1\textwidth}{!}{
        \begin{tabular}{cc}
            \hline
            Method & AUC \\
            \hline
            \rowcolor{gray!30}CoOp~\cite{zhou2022learning} & $71.44$ \\
            CoCoOp~\cite{zhou2022conditional} & $72.62$ \\
            \rowcolor{gray!30}FPL & $\mathbf{80.75}$ \\
            \hline
        \end{tabular}}
        \subcaption{\vspace{-3mm}}
        \label{tab:subtable2}
    \end{subtable}
    \caption{Comparisons to (a) existing CLIP-based forgery detection methods and (b) prompt learning methods.
    The performance is reported on FF++(c40).}
\end{table}
\Section{Conclusion}
In this work, we introduce M2F2-Det for interpretable deepfake detection.
Specifically, M2F2-Det uses the Forgery Prompt Learning to adapt CLIP's multi-modal learning ability for deepfake detection.
Then, an efficient Bridge Adapter connects the CLIP image encoder with a dedicated deepfake detection network, yielding more robust and effective visual representations for detection and seamlessly integrating with an LLM for enhanced interpretability. 
{
    \small
    \bibliographystyle{ieeenat_fullname}
    \bibliography{main}
}


\end{document}